\documentclass[conference]{IEEEtran}
\usepackage[T1]{fontenc}
\IEEEoverridecommandlockouts
\usepackage{amsmath,amssymb,amsfonts}
\usepackage{algorithmic}
\usepackage{graphicx}
\usepackage{textcomp}
\usepackage{xcolor}
\usepackage{url}
\usepackage{acro}
\usepackage{fancyvrb}
\usepackage[style=ieee]{biblatex}
\usepackage{xcolor,colortbl}
\usepackage{booktabs}
\usepackage{multirow}
\usepackage{makecell}
\usepackage{mathtools}
\usepackage{hyperref}
\usepackage{cleveref}
\usepackage[ruled, vlined, linesnumbered]{algorithm2e}
\newcommand{\revtext}[1]{\textcolor{black}{#1}}
\DeclareAcronym{bt}{
  short = BT,
  long  = Behavior Tree 
}

\DeclareAcronym{ros}{
  short = ROS,
  long  = Robot Operating System
}

\DeclareAcronym{rsass}{
  short = RSASS,
  long = Robotics Software Architecture-based Self-adaptive Systems
}

\DeclareAcronym{mapek}{
  short = MAPE-K,
  long  = Monitor-Analyze-Plan-Execute over Knowledge 
}

\DeclareAcronym{qos}{
  short = QoS,
  long  = Quality-of-Service
}

\DeclareAcronym{dsr}{
  short = DSR,
  long  = Deep State Representation
}
\DeclareAcronym{dsl}{
  short = DSL,
  long  = Domain-Specific Language 
}

\DeclareAcronym{ml}{
  short = ML,
  long  = Machine Learning
}

\DeclareAcronym{llm}{
  short = LLM,
  long  = Large Language Model 
}
\DeclareAcronym{iou}{
  short = IoU,
  long  = Intersection over Union 
}
\DeclareAcronym{uav}{
  short = UAV,
  long  = Unmanned Aerial Vehicle 
}

\DeclareAcronym{pddl}{
  short = PDDL,
  long  = Planning Domain Definition Language
}

\DeclareAcronym{fsm}{
  short = FSM,
  long  = Finite State Machine 
}

\DeclareAcronym{gsn}{
  short = GSN,
  long  = Goal Structuring Notation
}

\DeclareAcronym{mbd}{
  short = MBD,
  long  = Model-Based Diagnosis
}

\DeclareAcronym{rca}{
  short = RCA,
  long  = Root Cause Analysis 
}

\DeclareAcronym{ci}{
  short = CI,
  long  = Causal Inference
}
\def\BibTeX{{\rm B\kern-.05em{\sc i\kern-.025em b}\kern-.08em
    T\kern-.1667em\lower.7ex\hbox{E}\kern-.125emX}}

\begin{document}

\title{Who Is Responsible? Self-Adaptation Under Multiple Concurrent Failures With Unknown Faults in Complex Robotic Systems}

\author{\IEEEauthorblockN{Andreas Wiedholz$^{1}$,  Rafael Paintner$^{2}$, Alwin Hoffmann$^{1}$, Tobias Huber$^{1,3}$} 
\url{firstname.lastname@xitaso.com}, \url{rafael.paintner@dlr.de}, \url{tobias.huber@thi.de} \\
\textit{$^{1}$XITASO GmbH, Augsburg, Germany} \\
\textit{$^{2}$German Aerospace Center, Institute of Flight systems, Brunswick, Germany} \\
\textit{$^{3}$Technical university of Applied Sciences Ingolstadt, Germany} 

}

\maketitle
\begin{figure*}[h]
    \centering
    \includegraphics[width=\linewidth]{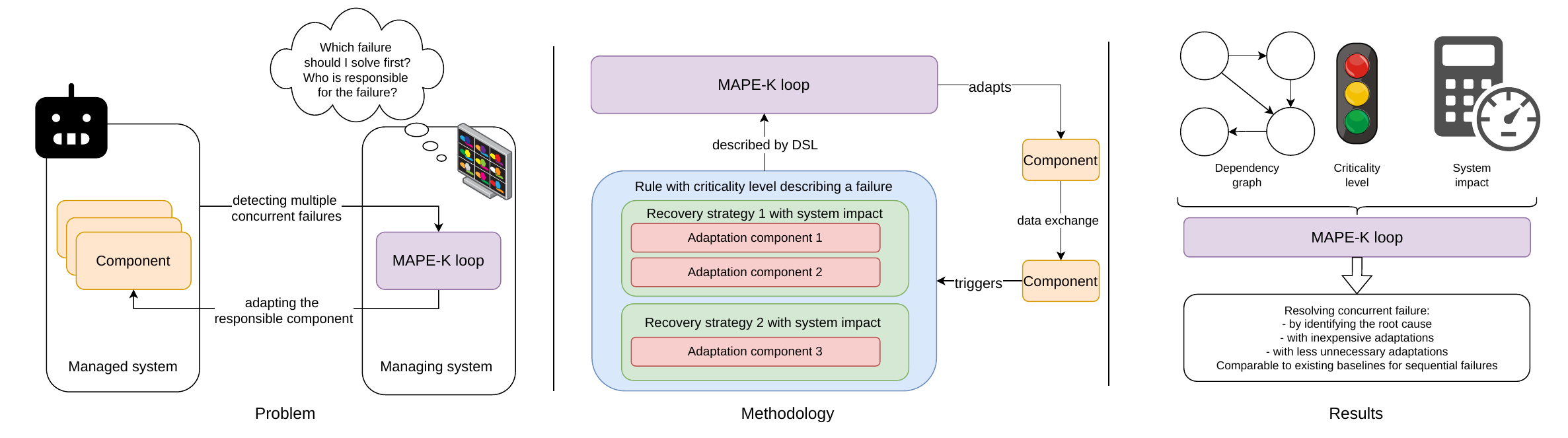}
    \caption{Overview over our DSL-driven approach: A MAPE-K loop that detects, ranks and resolves concurrent failures.}
    \label{fig:teaser-figure}
\end{figure*}
\begin{abstract}
Robotic systems increasingly operate in dynamic, unpredictable environments, where tightly coupled sensors and software modules increase the probability of a single failure cascading across components.
Therefore, multiple strategies can be plausible to resolve the underlying fault.
Most existing self-adaptive approaches that have been applied to robotics assume predefined one-to-one failure-to-adaptation mappings.
We present a ROS2-based self-adaptation approach building upon MAPE‑K that addresses (1) multiple simultaneous failures with differing criticality, (2) cascading failures across components, and (3) multiple plausible resolving strategies per detected failure. 
Central to our approach is an adaptation rule set which lets designers specify failure patterns, assign criticality levels, and enumerate multiple plausible adaptation strategies.
This rule set, combined with an automatically extracted live dependency graph, enables lightweight root-cause analysis and strategy ranking to prioritize minimal and effective adaptations.
Our approach implements a lightweight self-optimizing component which learns estimated success probabilities of different strategies for each known failure.
Experiments on an underwater robot scenario and a perception use case show that our approach can identify root causes among concurrent failures, favors inexpensive adaptations, reduces unnecessary adaptations, and achieves performance comparable to existing baselines designed for sequential failures.
The code is publicly available\footnote{\url{https://doi.org/10.5281/zenodo.21333401}}.
\end{abstract}


\section{Introduction}
In recent years, the field of robotics has witnessed a significant shift from operating in structured environments to handling dynamic and even unpredictable settings.
Thus, robots increasingly need to dynamically adapt to runtime circumstances by e.g., reconfiguring, starting, or stopping software components.
To tackle this challenge, several recent works applied techniques from the field of self-adaptive systems to robotic applications \cite{silva_mros_2023, alberts_rebet_2024, filippone_handling_2024, cheng_ac-ros_2020, silva_rosa_2025}.
These approaches aim to automatically handle failures or environmental changes.
Most of these approaches rely on the commonly used \ac{mapek} feedback loop \cite{kephart_vision_2003}, in which a managing system monitors and adapts a managed system consisting of the robot software. 

Modern robotic systems are composed of tightly interconnected components that depend on continuous data exchange \cite{silva_suave_2023, imrie_aloft_2024, wiedholz_sunset_2026}.
For example, most autonomous robot or drone systems contain multiple sensors like RGB and depth cameras, sensor fusion modules, and machine learning models responsible for perception tasks.
As a result, anomalous behavior in one component can easily cascade across the system, manifesting as multiple simultaneous or system-level failures \cite{timperleyROBUST221Bugs2024}.

While in some robotic systems the goal of a managing system is to protect the robot, i.e., returning it home safely \cite{imrie_aloft_2024}, we focus in this work on operating scenarios that expect robots to complete their missions despite failures that may occur and which should be corrected on the fly \cite{wiedholz_sunset_2026, silva_suave_2023}. 
With our approach, we want to solve the research question:
\textbf{How can multiple concurrent failures be resolved effectively in robotic systems at runtime?}
Here, we define resolving multiple concurrent failures as efficient if we are able to resolve these failures with as few adaptations in the system as possible.

In general, self-adaptive robotic approaches focused on sequential failures \cite{silva_mros_2023, alberts_rebet_2024, filippone_handling_2024} since the used exemplars are built for this aspect \cite{silva_suave_2023, askarpour_robomax_2021}.
Outside of the robotic domain, some self-adaptivity methods already tackle these problems \cite{vogel_language_2012, ghahremani_improving_2019,cailliau_runtime_2019}. 
However, applying these methods to the robotics domain is not straightforward for three main reasons.
First, translating the concepts to robotic systems like the \ac{ros} \,---\, a well-known middleware in robotics \cite{macenski_robot_2022} \,---\, is labor-intensive since such frameworks impose their own constraints.
Second, existing algorithms and implementations do not cater to the characteristics of the robotics domain, such as tight coupling to the physical world or introducing uncertainties and safety concerns.
Third, a large part of the code is not open source.

To address the aforementioned gaps, we adapt several solutions from self-adaptive systems outside of the robotic domain  \cite{vogel_language_2012, ghahremani_improving_2019} and apply them, to the best of our knowledge, for the first time to robotics by implementing them in a \ac{ros}-based system (see \Cref{fig:teaser-figure}).

The knowledge base in our approach consists of one static part defined at design time and one dynamic part, which is constantly updated at runtime.
For the static part, we create a \ac{dsl} that is inspired by Stitch \cite{CHENG20122860} and specifically designed for the needs of \ac{rsass} \cite{alberts_software_2025}.
It describes failures and mitigation strategies in a rule-based manner and requires design-time anticipation of failures that might occur during runtime.
Actually, for each anticipated failure, our \ac{dsl} is able to define multiple strategies to adapt different faulty components in the system that lead to the same failure.
The dynamic part of our approach creates a live dependency graph of the managed system, which allows us to consider dependencies in the managed system and resolve a failure after identifying its root cause.
We can prioritize which failure should be resolved first since our \ac{dsl} also allows us to categorize anticipated failures into three criticality levels.


We empirically validate that our approach can solve the aforementioned problems of current self-adaptive robotics in experiments with two robotic environments: (1) the commonly used benchmark SUAVE \cite{silva_suave_2023} and (2) SUNSET \cite{wiedholz_sunset_2026}.
Since \ac{ros} and \acp{bt} are commonly used in robotics \cite{colledanchise_behavior_2018, ghzouli_behavior_2023}, we implement our approach for the evaluation in \ac{ros} and the widely used BehaviorTree.CPP framework\footnote{https://www.behaviortree.dev/} ensuring easy access for the robotics community.

Our contributions can be summarized as follows:
\begin{itemize}
    \item An open-source self-adaptation approach for \ac{ros}-based systems performing \revtext{lightweight dependency-based} \ac{rca} to handle multiple concurrent failures with different criticality levels.
    \item A \ac{dsl} that enables robotics developers to describe the observable effects of runtime failures and different strategies to resolve these failures.
    \item A detailed two-fold evaluation including an ablation study investigating which information helps to choose the correct strategy to resolve concurrent failures in an efficient way.
\end{itemize}

\section{Related work}

\subsection{Multiple Concurrent Failure Resolving}
We first describe approaches that have not been applied to the robotics domain, while implementing similar ideas.
\textcite{vogel_language_2012} present an approach that deals with multiple concurrent faults that warrant adaptations with different performance-related impacts.
They differentiate between the goals of an adaptation, namely self-optimizing and self-repairing adaptations and implement them in sequential feedback loops. 
We implement both of these goals in one feedback loop with the ability to handle multiple failures concurrently instead of sequentially while also performing a \revtext{simple} \ac{rca}.

\revtext{Another goal-driven approach is presented by \textcite{cailliau_runtime_2019}. 
At design time, goal-oriented requirements engineering is required to identify failures preventing a goal from being reached. 
Similar to our approach, experts provide an initial belief of which failures prevent a goal from being reached, which can also be updated during runtime. 
Our approach differs as we perform a lightweight \ac{rca} to identify components that may be responsible for concurrent failures.}

\textcite{ghahremani_improving_2019} implement a rule-based and utility-driven approach, i.e., they choose a strategy to recover from a failure based on the estimated impact of the adaptation on the utility of the system.
The focus is on the goals of the system (e.g., avoid downtime) and the impact a strategy has on these goals.
Similarly, we prioritize strategies with high probability of avoiding downtime. 
In contrast, we perform a dependency-based \ac{rca} to derive the minimum set of adaptations, avoiding potentially excessive runtime actions. 

\revtext{Since \ac{ml} gains more attention in self-adaptive systems, rule-based approaches, which are still central in safety-critical scenarios, also became the basis for \ac{ml}-based approaches \cite{buresGeneratingAdaptationRulespecific2023}. 
\textcite{buresGeneratingAdaptationRulespecific2023} train a neural network based on existing rules to maintain the domain knowledge that otherwise might get lost during training without these rules. }

We extend the idea presented by \textcite{vogel_language_2012} since we consider the dependencies between the components of the managed system and executing self-optimizing adaptations if no component is involved in a critical failure.
Additionally, we use a rule-based approach similar to \textcite{ghahremani_improving_2019} and focus on the impact of strategies on the specific goal of avoiding downtime.
By considering the dependencies between software components, we extend the idea of categorizing adaptations by their criticality level in order to minimize the number of performed adaptations aiming for minimal system downtime.

\subsection{Robotic self-adaptation approaches}

There have been a number of approaches that apply self-adaptation principles to robotic systems. 
Most of them focus on self-healing scenarios that operate on the basis of specific robotic missions, either scenarios specifically designed for their evaluation \cite{cheng_ac-ros_2020, alberts_rebet_2024} or on benchmarks designed from the self-adaptation community \cite{silva_mros_2023, silva_rosa_2025}.

With AC-ROS \textcite{cheng_ac-ros_2020} evaluate \ac{gsn} models during runtime to monitor whether the software system can operate in the desired way. 
For this purpose, they describe multiple goals of the robot's behavior together with constraints to fulfill these goals, e.g., if all necessary sensors are available or a high enough battery level is given.
For violated constraints, they predefine adaptation plans to resolve each issue.
While AC-ROS has goals to fulfill rather than failures to resolve, they do not prioritize between the different goals in case multiple goals are violated at the same time.

The authors of ReBeT \cite{alberts_rebet_2024} introduce extensions of the BehaviorTree.CPP framework that help in applying task- and architecture-based adaptations in a robotic system.
They use \ac{bt} decorators for the \textit{Analyzing} and \textit{Planning} in the MAPE-K loop.
For specific \ac{bt} actions executed in the \ac{bt} they can adapt the respective \ac{bt} action node during runtime based on the current state of the managed system.
In ReBeT, the managing system is integrated into the managed system which is also restricted to being a \ac{bt}.
Therefore, when applying ReBeT to a new use case, the managed system has to be adapted, i.e., new \ac{bt} decorators have to be implemented.
This is different from our approach, in which only the rules describing how the managing system should resolve failures have to be created.
The code of the managing system does not have to be changed.

MROS (Metacontrol for ROS Systems), originally introduced in \cite{bozhinoski_mros_2022} and provided as a framework by \textcite{silva_mros_2023} integrates ontologies into the \ac{mapek} loop. 
Their ontology combines the \textit{Analyzing} and \textit{Planning}, which provides adaptations to execute based on ontological reasoning on possible adaptations and the current state of the managed system.
When applying this to a new use case, a new ontology has to be defined.
Although MROS is effective in resolving failures, it neither handles multiple failures at the same time nor considers the impact of multiple possible adaptations to resolve a failure.

ROSA \cite{silva_rosa_2025} uses a conceptual data model to describe the architecture of the managed system. 
It contains components, constraints, relationships, and adaptation plans.
Based on this architectural knowledge, different configurations of the components can be applied.
The selection of these configurations is performed by a reasoning engine that is built in TypeDB, a database that supports reasoning over a knowledge base.
This knowledge base includes rules describing how to react to certain constraint violations.
Similar to our approach, ROSA selects the components to adapt by priority of their robot system goal configurations and the dependencies between components.
Meanwhile, they do not consider which impact an adaptation has on the system's availability, which can lead to higher system down times. 

All of the above approaches assume that for every failure, one predefined strategy resolves the fault and puts the system back into an operational state. 
Our approach includes strategies to resolve not only failures with different criticality levels but also in situations when the responsible \ac{ros} node and therefore the strategy to resolve the failure is not predefined.
By categorizing failures into criticality levels and considering dependencies in the system, our approach is able to prioritize which failures should be resolved first and which strategy is the one with the highest probability of success.

\subsection{Root cause analysis}
Identifying the source fault of a failure requires a root cause analysis. Therefore, this section discusses several related works from this area. 
In robotic \ac{mbd}, domain knowledge of the managed system is embedded in complex mathematical models that estimate the current state of the system \cite{hasan_model-based_2023}.
These models are typically based on \ac{ml}, statistical analysis, or designed by experts \cite{sabry_review_2024}.
While \ac{mbd} can localize faults with a high accuracy, the system still needs to find an optimal way to resolve the failure.
Our work does not try to embed domain knowledge in detailed mathematical models.
Instead, it utilizes the existing system architecture of the managed system with the additional expert knowledge about different strategies to resolve common failures on an architectural level.

In recent years, \ac{ci} has attracted attention in IoT~\cite{xin_causalrca_2023} and robotic research.
The goal in robotic \ac{ci} is to detect dependencies between system states~\cite{hellstrom_relevance_2021} or events that lead to a failure \cite{diehl_causal-based_2023}.
Even though this method allows building up models that represent dependencies inside the managed system, they usually require either large training data or extensive domain knowledge \cite{hellstrom_relevance_2021}.
In robotics applications that are using \ac{ros}, the ros\_rqt graph presents dependencies on an architectural level, and therefore, they do not need to be learned.

In contrast to our approach, the above-mentioned approaches only focus on the diagnosis part and not on adapting the managed system. 
Furthermore, \ac{mbd} and \ac{ci} are too computationally expensive for robotic use cases with limited hardware resources and real-time requirements.
Our work proposes a lightweight approach that only requires architectural knowledge embedded in a \ac{dsl} instead of complex mathematical or \ac{ml} models.

\section{Problem}
\label{sec:problem-formulation}
This section defines the problem of multiple concurrent failures in robotic systems with unknown sources.
We implement a practical exemplar for this problem in SUNSET\cite{wiedholz_sunset_2026}, which simulates a sensor fusion-based perception pipeline for a drone.
The drone continuously segments its environment, and therefore, all components in the segmentation pipeline are dependent on constant data exchange.
For instance, if the drone’s camera fails, downstream modules such as sensor fusion and the perception model can no longer produce data, leading to an outage of several components simultaneously.
Similarly, the same downstream failure can be caused by different faults:
E.g., when the perception model becomes uncertain about its predictions, potential reasons include degraded sensor data or misalignment during sensor fusion requiring fundamentally different adaptation strategies.

Let $C = \{c_1, ...,c_n\}$ be the set of components the managed system consists of.
For $c \in C$, $F_c = \{f_{c,1},\ldots f_{c,m_c}\}$ represents perturbations containing software internal faults that may be present in $c$ during runtime or environmental disturbances monitored by $c$.
Therefore $F = \cup \{F_c | c\in C\}$ contains all possible faults in the managed system that \revtext{are known at design time and} result in failures.
We define a set of design-time known failures $T = \{t_1,\ldots,t_k\}$ with each $t_i \in T$ describing a concrete event or measurement caused by a fault. 
Furthermore, $K_\text{system}$ contains the knowledge about the state of the managed system and the environment that can be used to evaluate if a failure is present.
Lastly, we define an influence relation $\Phi \subseteq F \times T$ where $(f_c,t_i) \in \Phi$ indicates that a fault $f_c \in F$ results in a detectable failure $t_i$.

At runtime, the subset $F^t \subseteq F$ represents the faults that are actually active at time $t$. 
We assume $F$ is known (all possible faults) but $F^t$ is unknown and contains multiple faults ($|F^t| > 1$). 
Furthermore, the knowledge $K_\text{system}^t$ represents the updated system information at a time $t$.
While we can detect currently present failures $T^t \subseteq T$ considering $K_\text{system}^t$, we do not know $F^t$ since $\Phi$ does not describe a one-to-one relation.
For a given failure $t_1\in T$ there can be several faults that could have caused it: e.g. $f_1, f_2  \in F$ such that $(f_1,t_1) \in \Phi \land (f_2,t_1) \in\Phi$.
Similarly, given a fault $f_1\in F$, there can be several failures that it causes since $f_1$ propagates through multiple components: e.g., $t_1,t_2 \in T$ such that $(f_1,t_1) \in \Phi \land (f_1,t_2) \in\Phi$.

Therefore, we derive the following problem:
Given the design-time knowledge ($C, F, T, K_\text{system}, \Phi$) we want to infer at each time $t$ the set of faults $F^t$ and which faults $f_i \in F^t$ should be solved first to effectively reach the goal of no detectable failures $|T^t| = 0$.

\section{Approach}
\label{ssec::system_design}

The target architecture of our approach consists of a managed system performing the task and a managing system that monitors and adapts this managed system.
Our work assumes that the managed system is already present and focuses on the design of the managing system.

The managed system consists of the robotic system, including all algorithmic, hardware, and controlling components such as \acp{bt} \cite{colledanchise_behavior_2018}, \acp{fsm} \cite{ghzouli_behavior_2023}, or reinforcement learning agents \cite{maroto-gomez_systematic_2023}. 
We assume that the managed system can autonomously perform its mission without the presence of faults.
Moreover, we assume that the managed system is built from components that periodically communicate with each other, as it is common practice in robotic use cases.
In \Cref{sec:evaluation}, we describe two examples of such systems that we used for our evaluation.

For the managing system, we build upon the well-known \ac{mapek} loop \cite{kephart_vision_2003} that is executed periodically.
We designed the managing system to be solely responsible for failure resolution, i.e., identifying the fault and adapting the system accordingly.
The managing system operates independently of the managed system, similar to MROS \cite{bozhinoski_mros_2022} or ROSA \cite{silva_rosa_2025}.
The managed system exposes interfaces used by the managing system to resolve failures in the relevant components.


We build upon the idea presented in ROSA \cite{silva_rosa_2025} by using a set of \textbf{self-adaptation rules} $R$ to describe when and how to resolve a failure.
The rules are part of the knowledge in the \ac{mapek} loop \revtext{and require domain knowledge about architectural and behavioral characteristics of the managed system}.
In our approach, we extend the idea by the following:
(1) Each rule $r\in R$ can have multiple strategies $S_r$ that can be used to solve different faults that lead to the same failure.
(2) Each rule has a criticality level $l_r$ to prioritize in case multiple failures are detected simultaneously.
(3) Based on the criticality level $l_r$ of a rule $r$ and the dependencies between the components in the managed subsystem, we can select multiple strategies $s\in S_r$ at once as long as they do not interfere with each other (see \Cref{ssec:metho_managing_system}).

\subsection{Self-Adaptation Rules}
\label{sec:rules_description}
\begin{figure}
    \centering
    \includegraphics[width=\columnwidth]{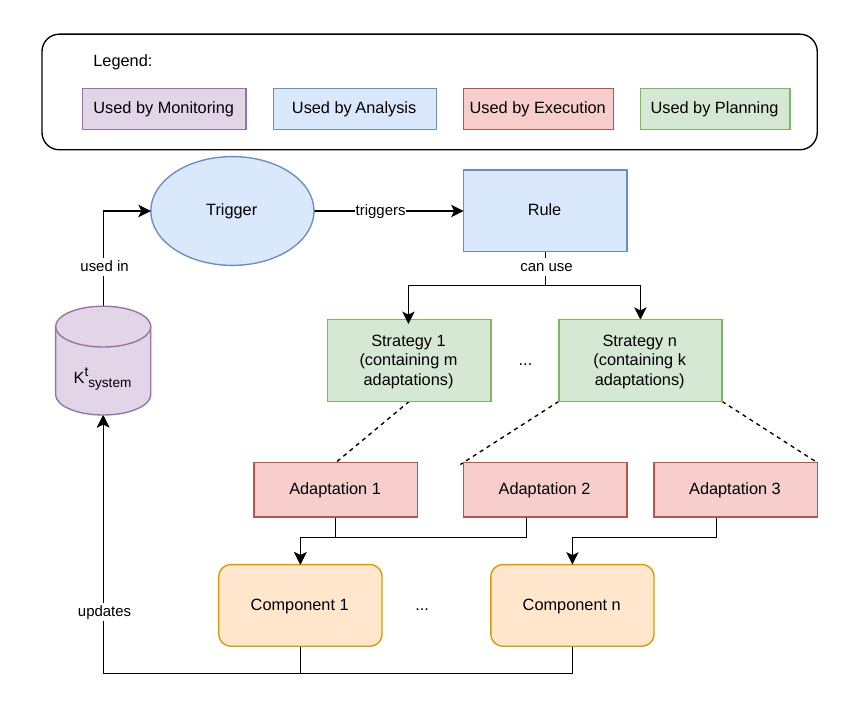}
    \caption{Relationship between rules, strategies, adaptations and ROS nodes in the managed system.}
    \label{fig:rule_strategy}
\end{figure}
Before we describe how our proposed approach implements the MAPE-K loop, this subsection introduces how we define our self-adaptation rules, as well as the \ac{dsl} that allows designers to easily create new rules.
As depicted in \Cref{fig:rule_strategy}, the main components of each rule $r\in R$ are the failure $t_r \in T$, a set $S_r$ of strategies that resolve the failure, and the criticality level $l_r \in \{\text{\texttt{NON\_CRITICAL}, \texttt{WARNING}, \texttt{ERROR}}\}$.

A rule $r \in R$ describes a failure $t_r$ in the managed system with a trigger that is measurable through $K^t_\text{system}$ and potential strategies to resolve the failure.
To resolve a detected failure $t_r$, each rule defines a set of \textbf{strategies} $S_r$.
Each strategy $s \in S_r$ aims to adapt at least one $c \in C$ that could cause the failure $t_r \in T$ detected its respective the trigger.
To better estimate which strategy to choose first, the system designer has to provide an estimated success probability $P_s$ for each strategy $s$. 

To resolve the failure, each strategy $s$ in turn consists of a set of \textbf{adaptations} $A_s$. 
Each adaptation $a\in A_s$ affects a single component in the managed system.
Based on recent analyses of adaptation mechanisms in \ac{rsass} \cite{alberts_software_2025}, we designed our approach to support four types of adaptation:
\begin{itemize}
    \item Re-parametrization of a component $c$, i.e. changing parameter values that are used in the components tasks.
    \item Adaptation of the communication channel, e.g. changing the input of a component.
    \item Adding/removing components.
    \item Redeployment of components.
\end{itemize}

Each adaptation $a$ is assigned with an adaptation overhead $i_{\text{exec},a}$ which equals the number of cycles in the managing system it is expected to take to resolve the failure (e.g., based on prior measurements). 
Note, that this can be different for the same type of adaptation for different components.
With the adaptation overhead $i_{\text{exec},a}$, we calculate the \textbf{system impact} of a strategy $i_{\text{exec},s}$ by a worst case analysis: $i_{\text{exec},s} \coloneq \max_{a \in A_s}(i_{\text{exec},a})$.

To facilitate the prioritization of critical rules within our system, we define the following three \textbf{criticality levels} of failures:
\begin{itemize}
    \item \texttt{NON\_CRITICAL}: Adaptation for a subsystem of the managed system that is mission specific. The executed adaptation is part of the planned mission and the affected subsystem remains fully operable, thus no inferences with other adaptations occur. 
    \item \texttt{WARNING}: A component publishes data that is usable but should be improved. A misconfiguration of the managed system is more likely to be the reason than changes in the environment, e.g. a recalibration between sensors.
    \item \texttt{ERROR}: A part of the managed system isn't operable anymore, e.g., it cannot publish new data. The reason for that can be a sensor outage or missing input data without which a component cannot further produce new outputs itself, e.g. an outage of a sensor.
\end{itemize}

\subsubsection{Our Domain Specific Language}

\begin{figure} \centering \begin{Verbatim}[fontsize=\tiny] 
1 RULE <name> 
2   POLICIES <crit_level>
3   TRIGGER <expression>
4     STRATEGY <name> <success probability> <initial belief>
5       ADAPTATION <component> <type> <parameters> <system impact>
6       ADAPTATION <component> <type> <parameters> <system impact>
7     STRATEGY <name> <success probability> <initial belief>
8       ADAPTATION <component> <type> <parameters> <system impact>
\end{Verbatim} 
\caption{A generic rule in the syntax of our \ac{dsl}.} 
\label{fig:syntax-dsl}
\end{figure}

To enable system designers to create new rules easily, we developed a \ac{dsl} that allows them to express failures, strategies, and adaptations concisely.
\Cref{fig:syntax-dsl} shows the generic syntax of a rule.

\begin{figure} \centering \begin{Verbatim}[fontsize=\tiny] 
1 RULE SegmentationBad
2   POLICIES WARNING 
3   TRIGGER segmentation_entropy > 0.06
4     STRATEGY recalibration 80 50
5       ADAPTATION fusion set_parameter recalibration true 2
6     STRATEGY enhancement_activate 10 50
7       ADAPTATION image_enhancement activate 5
8       ADAPTATION fusion change_communication camera_input rgb_enhanced 2
9     STRATEGY enhancement_deactivate 10 10
10      ADAPTATION image_enhancement deactivate 5
11      ADAPTATION fusion change_communication camera_input rgb_raw 2
\end{Verbatim} 
\caption{A specific rule in the syntax of our \ac{dsl} for the SUNSET \cite{wiedholz_sunset_2026} use case.}
\label{fig:specific_syntax-dsl}
\end{figure}

Basic parameters \,---\, such as rule name, the criticality level $l_r$, strategy names, strategy success probabilities, etc. \,---\, are defined with simple data types like strings and floats.
The TRIGGER describing how a failure $t_r$  will be detected at runtime based on the current $K_{\text{system}}^t$ is defined by a C++ style notation expression that can handle constant values as well as system variables of different data types. 
These expressions are brought into Reverse Polish Notation by a shunting yard algorithm \cite{Dijkstra1963} and stored as a tree structure, which allows us to evaluate expressions in a recursive way.
Each ADAPTATION $a$ specifies which component in the managed system it adapts, which type of adaptation it implements, and the adaptation overhead as defined in \Cref{sec:rules_description}.

Depending on the type of adaptation, the parameters of the ADAPTATION are different. 
Redeployment and activation/deactivation only need the name of the component.
Communication change is specified by the names of the component and the new input channel. 
Re-parametrization is defined by the name of the parameter to adapt, followed by the new value of the parameter as a C++ style notation expression.
\Cref{fig:specific_syntax-dsl} shows an exemplary rule for SUNSET~\cite{wiedholz_sunset_2026} in our \ac{dsl}. 
This specific rule \revtext{with a \texttt{WARNING} criticality level (line 2) detects the trigger $t_r$} when the image segmentation quality drops \revtext{(line 3)}. 
To solve this failure, \revtext{three strategies $s_{1-3} \in S_r$ are defined.} Either a recalibration in the sensor fusion is performed \revtext{(line 4)}, the image enhancement is activated \revtext{(line 6)}, or the image enhancement is deactivated \revtext{(line 9)}. 

\subsection{Managing system}
\label{ssec:metho_managing_system}

In this section, we describe how our approach implements the individual parts of the MAPE-K loop and our light-weight learning mechanism.

\subsubsection{Knowledge base}
Our \textit{Knowledge base} has four components: The live \textbf{dependency graph} of the managed system, the \textbf{system state information} $K^t_{\text{system}}$ at any given time $t$, the \textbf{set of adaptation rules $R$} including the failures $T$ in the system caused by the $F$, and the \textbf{strategies to resolve the failures} \revtext{anticipated at} design time (see Section \ref{sec:rules_description}).
The dependency graph is automatically created and updated by our \textit{Monitoring} reflecting the current architecture of the managed system, including the communication channels that we consider as dependencies between components.
We use the dependency graph to detect ripple effects, i.e., a fault in one component leads to failures in other components.

$K_\text{system}$ stores the information that is inherently present in the managed system and estimations we can automatically extract from established communications.
Examples for this information are operative health parameters, e.g., frequencies of published topics, and mission-related information, e.g., entropy of an AI model or environmental conditions.
To store $K_\text{system}^t$, we use a Key-Value storage with keys that can be used in the triggers to detect a failure $t_i \in T$.

\subsubsection{Monitoring}
Our \textit{Monitoring} has two responsibilities. 
First, it creates and updates $K_\text{system}^t$.
The second responsibility of our proposed \textit{Monitoring} is the creation of the dependency graph of the managed system by mapping the currently present components $C_t \subseteq C$ and their established communication ways into our dependency graph.
Thus, new or removed dependencies are always considered in the \textit{Planning}.

\subsubsection{Analyzing}
The \textit{Analyzing} performs two tasks.
Firstly, failures are detected by evaluating the triggers in the adaptation rule set $R$.
Secondly, in case a strategy has been selected for execution in previous cycles by the \textit{Planning}, it evaluates which of these strategies resolved a failure.

In each cycle, we update $T^t$ by evaluating for all adaptation rules $r\in R$ if a failure $t_r$ detected by its trigger is present by using the latest $K_{\text{system}}^t$. 
If a new failure is detected, the rule is handed to the \textit{Planning} for further processing.
Furthermore, in the dependency graph, all components affiliated with said rule $r$ are set to the rule's criticality level $l_r$. 
After a strategy $s$ is selected for execution by the \textit{Planning}, the \textit{Analyzing} waits the number of cycles defined in $i_{\text{exec},s}$ and continuously checks if the failure has been resolved by evaluating the trigger.
If not, we assume that the recently executed strategy does not apply to the underlying fault, and the rule is handed to the \textit{Planning} again. 
Additionally, the \textit{Planning} receives the information that the executed strategy did not resolve the failure to guide further decision-making.
On the contrary, if the failure is resolved, we consider the last selected strategy successful. 

\subsubsection{Planning}
\label{sssec:planning}
The \textit{Planning} is responsible for deciding which strategies $s \in S_r$ should be used for each of the triggered rules $r \in R$.
The planning algorithm performs a probabilistic \revtext{dependency-based} \ac{rca} based on three factors:

(1) Considering \textbf{dependencies} between components: The goal of using this information is to find the root cause in case a fault in component $c_n$ generates failures in other components $c_m, ..., c_{m+k}$ depending on $c_n$.
(2) Considering the \textbf{criticality level}: The goal of using this information is to first resolve failures with a high criticality level, which may lead to failures with a lower criticality level resolving themselves. This creates a prioritization dependency.
(3) Considering the \textbf{system impact}  $i_{\text{exec},s}$ of different strategies  $s\in S_r$ for the given rule $r$: The goal of using this information is to choose a strategy that promises to resolve a failure faster than other strategies.


The first step of our \textit{Planning} is to combine all strategies that belong to a triggered rule with criticality level ERROR in a set $S_\texttt{ERROR}$.
Analogously, we create $S_\texttt{WARNING}$ and $S_\texttt{NOT\_CRITICAL}$.
For each criticality level, we sort all of the strategies by the cost of the strategy.
We define the \textbf{cost $\text{Cost}_s$ of a strategy} $s$ according to \Cref{eq:cost}:
\begin{equation}
    \text{Cost}_s = (100 - P_{s})/100 + (i_{\text{exec},s} / i_\text{max})
    \label{eq:cost}
\end{equation}
where \( P_{s} \) and \(i_{\text{exec},s}\) are the success probability of strategy $s$ (expressed as a percentage) and the strategy system impact to execute $s$ as defined in \Cref{sec:rules_description}.
The strategy system impact $i_{\text{exec},s}$ is normalized with the highest strategy impact $i_\text{max}$ defined in the rules for the specific use case.

After categorizing the strategies by criticality level, the planning algorithm has to consider which strategies conflict with each other.
Since our tooling detects conflicts inside a strategy already during design time (see \Cref{sec:reusability}), our planning algorithm only needs to identify invalid adaptations and interferences between strategies.
To identify invalid adaptations, the planning algorithm compares expected results with the current system's state, e.g., it checks if an adaptation adds a node to the system that is already active.
In case of multiple detected failures, we want to select as many strategies concurrently as possible with the constraint of one strategy per failure.
Strategies that can be executed simultaneously are interference-free.
To select as many interference-free strategies as possible, we propose Algorithm \ref{alg:planning}.

For each criticality level, the algorithm iterates over the sorted strategies and selects the strategies that fulfill the following requirements:
\begin{enumerate}
    \item The strategy has not been tried yet. (Line 5)
    \item The strategy is valid considering the current state of the system, e.g., it adds a component to the system that is currently not present. (Line 7)
    \item The strategy doesn't affect a component that is already being adapted. (Line 10)
    \item The strategy doesn't affect a component already being affected by a strategy selected previously in the current cycle. (Line 12)
    \item The strategy doesn't affect a component that is dependent on another component that is potentially affected by a triggered rule with the same or a higher criticality level. (Line 14)
\end{enumerate}
All of this information is obtained from our dependency graph and the adaptation rules, which are maintained by the \textit{Monitoring} and the \textit{Analyzing} respectively.
After selecting a strategy, we store the affected components for the current cycle in order to not select another strategy affecting the same nodes.
\begin{algorithm}
\small
\caption{Strategy selection}
\label{alg:planning}
\SetKwProg{generate}{Function}{}{end}
\SetNlSty{textbf}{}{}         
\SetNlSkip{1em}               
\DontPrintSemicolon          
\generate{select strategy (criticality\_level, graph, detectedFailures)}{
    possibleStrategies = sortStrategiesByCost(detectedFailures)
    
    \revtext{plannedStrategies = list()}
    
     \ForAll{strategy in possibleStrategies}{
        \If{strategy.alreadyTried}{
            continue;
        }
        
        \If{strategy.invalid(graph)}{
            continue;
        }
        \ForAll{nodes in strategy.affectedNodes}{
            \If{node.currentlyAdapted}{
                 break;
            }
            \If{node.alreadyAffected}{
                 break;
            }
            \If{graph.relevantDependenciesPresent(node, criticality\_level) == True}{
                break;
            }
        }
        strategy.setAffectedNodes(True);
        
        \revtext{plannedStratgies.append(strategy)}
        
     }
     
    return \revtext{plannedStrategies;}
}
\end{algorithm}

\subsubsection{Execution}
The \textit{Execution} is responsible for adapting the managed system by implementing the adaptations $a$ for all strategies $s$.   
As described in \Cref{ssec::system_design} we consider four types of adaptation: reparameterization, communication change, addition/removal, and redeployment of a component.
In our approach, the components of the managed system offer a service that executes the component-specific adaptation and can be called, therefore, by the managing system.
Further, we assume that the parametrization of a component can trigger a component-specific routine in case the component is not able to detect the necessity of this routing, e.g., a new calibration between sensors in a sensor fusion component.
Since the implementation of concrete mechanisms for the adaptation depends on the used middleware, e.g., \ac{ros}, this will not be explained in more detail here.

\subsubsection{Success Rate Adaptation}
\label{sssec:learning}
In a real system, the true success rate of a strategy might differ from the initial assumption at design time. Furthermore, the underlying occurrence probability of faults may vary over time.

This is addressed by adapting the initially assumed success rate of each strategy after every execution attempt, such that it eventually converges towards the true success rate.
We assume the probability density of the success rate of each strategy $p_s$ to follow a Beta distribution with hyperparameters $\alpha$ and $\beta$: $p_{s,\text{prior}} = \text{Beta}(\alpha, \beta)$. 
Following Bayes' theorem, once a strategy has been selected, the probability density function of its success rate is updated to: $p_{s,\text{posterior}} = \text{Beta}(\alpha + 1, \beta)$, if the failure was resolved and to: $p_{s,\text{posterior}} = \text{Beta}(\alpha, \beta + 1)$, if it was not. 
In  \Cref{eq:cost} we use the mean, i.e., the expected value of the current probability distribution: $P_s = \mathbb{E}(p_s) = \frac{\alpha}{\alpha + \beta}$. 
Put simply, we log the result of each execution attempt, starting from a prior pseudo-count that represents our first initial belief, and update the success probability by calculating the percentage of successful execution attempts.\cite{Hoff_AFirstCourseinBayesianStatisticalMethods}

To derive the initial distribution, as seen in \Cref{fig:syntax-dsl}, the user has the opportunity to provide an initial belief about the true success rate $P_{s,{\text{init}}}$, as well as a confidence level about their belief $c_{p}$. 
The hyperparameters of the initial Beta function are derived as follows: $\alpha_\text{init} = c_{p} P_{s,\text{init}} $ and $\beta_\text{init} =c_{p} (1-P_{s,{\text{init}}})$. 
Thus, a high confidence about the initial belief leads to less influence of each update. 
Contrarily, a low confidence leads to a strong influence of success rate adaptations and possibly fluctuations during early runtime.

\section{Evaluation}
\label{sec:evaluation}

Based on the overarching research question "How can multiple concurrent failures be resolved effectively in robotic systems at runtime?" we want to answer the following research questions with our evaluation:

\begin{itemize}
    \item RQ1: To what extent can our approach manage concurrent, cascading failures in complex robotic systems?
    \item RQ2: How do the dependency graph, criticality level and system impact components of our approach affect its performance in a scenario with multiple concurrent failures?
\end{itemize}

Furthermore, we investigate a third research question (RQ3) regarding traditional self-adaptive robotic problems that do not involve multiple concurrent failures:
\begin{itemize}
    \item RQ3: How does our approach compare against the state-of-the-art for multiple sequential failures?
\end{itemize}

In order to answer these questions, we perform a two-fold evaluation based on the SUNSET \cite{wiedholz_sunset_2026} and the SUAVE~\cite{silva_suave_2023} exemplar.

\subsection{Implementation}
To evaluate our approach in a robotic context, we implement it in \ac{ros}2 humble.
This makes our approach accessible to the robotics community while it also allows us to evaluate it on two exemplars \cite{silva_suave_2023, wiedholz_sunset_2026}.
To implement the cycles in our \ac{mapek} loop, we integrate our approach in \acp{bt} in the well-established BehaviourTree.CPP framework.
The adaptations are implemented as service clients since both of the managed systems in the evaluation scenarios offer services that execute the requested adaptation.
We build upon the rqt\_graph in \ac{ros} for our dependency graph and enrich it with the information that our \textit{Planning} component bases its decisions for strategy selection on.
Since the rqt\_graph is already present in \ac{ros} systems, there is almost no computational overhead with building our dependency graph upon it.
Furthermore, we implement all failures and their mitigation strategies in the managed system in our \ac{dsl}.

\subsection{Multiple concurrent failures}
\label{ssec::evaluation_scenario}

For our first evaluation scenario, we use SUNSET presented in \cite{wiedholz_sunset_2026}.
This exemplar contains multiple concurrent failures with varying criticality levels, which we use without modifications.

SUNSET \cite{wiedholz_sunset_2026} describes a drone use case consisting of an RGB and depth camera, an image enhancement node, a sensor fusion node, and a semantic segmentation node.
The RGB and depth images are collected in the sensor fusion node and then jointly processed by the semantic segmentation. 
Additionally, the image enhancement node can be employed to remove potential artifacts from the RGB image. 
SUNSET introduces 10 faults causing five different failures that can be categorized into three different criticality levels.
The failures possible to simulate are outages of the camera (F1), sensor fusion (F2), or segmentation node (F3) due to different reasons; a decreased performance of the segmentation algorithm (F4) and a mildly blurred camera image (F5).
Each failure can be caused by multiple faults with different strategies to resolve them with.
The strategies contain all of the adaptations our system is capable of (redeploy, reactivate, communication change, reparametrization). 

For our test runs, we use the scenarios provided in the SUNSET exemplar which clearly define which faults are introduced into the system.
This ensures that the results in our experiments remain comparable.
Each experiment consists of 162 runs, which equals every scenario including one of the possible fault combinations being executed 9 times. 
\subsubsection{Ablation study}
First, we perform an ablation study demonstrating the effect of every component in our planning algorithm.
As described in \Cref{ssec:metho_managing_system}, the planning algorithm can be parametrized in order to consider the dependencies between \ac{ros} nodes, the different criticality levels, and the system impact of a strategy.
The goal when using a managing system may differ from application to application, e.g., different use cases might necessitate an optimization of the system's behavior with respect to the least amount of adaptations, the least downtime of the system, or the fastest reaction time to a failure.
For the ablation study, we set the initial $P_s$ for each strategy to the ground truth value as described in SUNSET.
We use the metrics defined in \cite{wiedholz_sunset_2026} to evaluate the impact of the different parameterizations of our planning algorithm.
The results for all eight possible planning algorithm parameterizations are shown in \Cref{tab:ablation-study}.
\begin{table}[ht]
\caption{Results of Ablation Study. The first three columns indicate the considered configuration, i.e. whether the dependency graph, the criticality level, or system impact are considered. The other cells show the mean and standard deviation of each metric.}
\label{tab:ablation-study}
\centering
\begin{tabular}{c@{\hspace{2pt}}c@{\hspace{2pt}}c|cccc}
\toprule
\makecell{D} & \makecell{C} & \makecell{S} & $\frac{f_\text{resolved}}{s_\text{executed}}$ & $t_\text{react}$[s]  & $t_\text{down}$[s] & $\text{\#rdply}_\text{u}$ \\
\midrule

x & x & x &                             0.75 $\pm$ 0.20 & 2.63 $\pm$ 1.76 & 4.90 $\pm$ 1.12 & 1.41 $\pm$ 1.11 \\
x & x & \checkmark &                    0.78 $\pm$ 0.19 & 2.16 $\pm$ 2.50 & 4.72 $\pm$ 2.73 & 0.56 $\pm$ 0.83 \\
x & \checkmark & x &                    0.74 $\pm$ 0.19 &  2.69 $\pm$ 1.81 & 5.37 $\pm$ 1.43 & 1.49 $\pm$ 1.18 \\
x & \checkmark & \checkmark &           0.79 $\pm$ 0.18 &  \textbf{1.95} $\pm$ 2.42 & 4.65 $\pm$ 2.83 & 0.49 $\pm$ 0.75 \\
\checkmark & x & x &                    0.83 $\pm$ 0.17 &  2.11 $\pm$ 1.62 & 5.17 $\pm$ 1.16 & 1.43 $\pm$ 1.13 \\
\checkmark & x & \checkmark &           0.81 $\pm$ 0.16 &  2.01 $\pm$ 2.42 & 4.51 $\pm$ 2.51 & 0.32 $\pm$ 0.67 \\
\checkmark & \checkmark & x &           \textbf{1.38} $\pm$ 0.47 &  2.79 $\pm$ 1.99 & 5.20 $\pm$ 1.65 & 0.81 $\pm$ 0.98 \\
\checkmark & \checkmark & \checkmark &  1.01 $\pm$ 0.33 &  1.97 $\pm$ 2.41 & \textbf{4.49} $\pm$ 2.49 & \textbf{0.18} $\pm$ 0.45 \\

\bottomrule
\end{tabular}
\end{table}

\subsubsection{Robustness analysis}
To show the effect of an incorrect estimation of the success probabilities $P_s$ for each strategy in the rule set, we perform a robustness analysis.
In this analysis, the initial $P_s$ are 20\% off the ideal estimations used in the ablation study.
We compare the results of this experiment with the best setting we found in \Cref{tab:ablation-study}.
\Cref{tab:robustness test} demonstrates that for the SUNSET use case, there is no significant difference ($p>0.05$) between ideal initial $P_s$ and misspecified ones.

\begin{table}[]
    \caption{Results of robustness analysis}
    \label{tab:robustness test}
    \centering
    \begin{tabular}{c|cccc}
    \toprule
    Initial $P_s$ &  $\frac{f_\text{resolved}}{s_\text{executed}}$ & $t_\text{react}$[s]  & $t_\text{down}$[s] & $\text{\#rdply}_\text{u}$\\
    \midrule
    Ideal & 1.01 $\pm$ 0.33 & 1.97 $\pm$ 2.41 & 4.49 $\pm$ 2.49 & 0.18 $\pm$ 0.45 \\
    \makecell{Mis- \\specified} & 0.98 $\pm$ 0.30 & 2.02 $\pm$ 2.52 & 4.48 $\pm$ 2.89 & 0.26 $\pm$ 0.59 \\
    \bottomrule
    \end{tabular}
\end{table}

\subsubsection{Learning mechanism}
\begin{figure}
    \centering
    \includegraphics[width=\columnwidth]{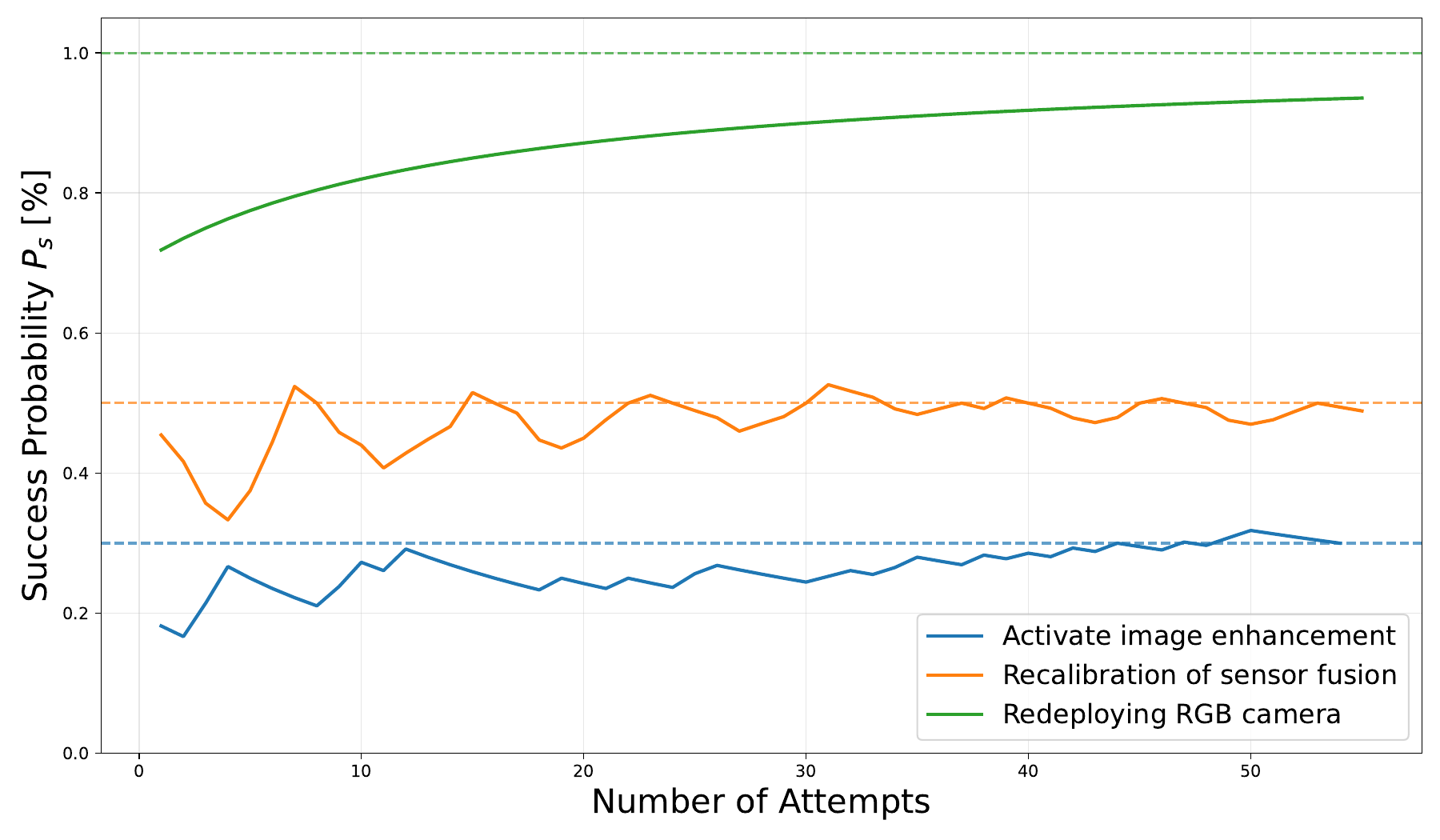}
    \caption{Estimation of $p_s$ to resolve F1 and F4 in SUNSET over time with the learning component enabled in our approach.}
    \label{fig:learning-result}
\end{figure}

In this experiment, we enable the learning mechanism described in \Cref{sssec:learning}.
The initial success probabilities are set as in the robustness analysis, and the number of experiments is the same as in the ablation study. As an initial belief, a value $c_p$ of 10 was chosen.
Overall 180 faults were introduced during this experiment, causing in total 257 failures.
\Cref{fig:learning-result} shows exemplarily for F1 and F4, that the learned $P_{s, \text{learned}}$ converge\revtext{s} to the \revtext{true success rate} $\revtext{P}_{s,\text{ideal}}$. 

\subsection{Multiple sequential failures}
\label{ssec:eval_single}

To compare our approach with the state of the art in resolving multiple sequential failures \cite{silva_mros_2023, silva_rosa_2025}, we apply our managing system to the SUAVE exemplar \cite{silva_suave_2023}. 
Since SUAVE uses the system modes \,---\, a different implementation of reparametrization and lifecycle changes than used in SUNSET \cite{wiedholz_sunset_2026}\,---\, presented by \textcite{nordmann_system_2021}, we extend our managing system, enabling it to use these system modes.
For this use case, the task is to react to failures in an underwater mission in which a robot is searching and inspecting a pipeline.
First, changes in the visibility of the water necessitate an adaptation of the search behavior (U1).
Second, a failure of the thrusters of the underwater robot necessitates performing a thruster recovery (U2).
In the extended SUAVE exemplar \cite{silva_rosa_2025}, there is a third failure, which is a low battery level that necessitates recharging the robot (U3).

\begin{table}[htb]
    \centering
    \caption{Mission results on SUAVE (100 runs each). U1: Low water visibility, U2: Thruster failure, U3: Low battery level. Results for \textit{None, BT, Metacontrol}, and \textit{ROSA} are taken from \textcite{silva_rosa_2025}.}
    \label{tab:comparison_sota}
    \setlength{\tabcolsep}{4pt}
    \begin{tabular}{@{}lcccccc@{}}
        \toprule
        \textbf{System}
            & \textbf{Search time [s]}
            & \textbf{Dist.\ insp.\ [m]}
            & \multicolumn{3}{c}{\textbf{React.\ time [s]}} \\
        \cmidrule(lr){4-6}
            & \textbf{Mean $\pm$ Std}
            & \textbf{Mean $\pm$ Std}
            & \textbf{U1} & \textbf{U2} & \textbf{U3} \\
        \midrule
        \rowcolor{gray!15}
        \multicolumn{6}{@{}l}{\textbf{SUAVE}} \\
        None        & $174.75 \pm 36.00$ & $33.20 \pm 13.49$ & N/A           & N/A           & N/A \\
        BT          & $84.09  \pm 26.41$ & $\mathbf{62.70 \pm 7.78}$  & $\mathbf{0.08}$ & $\mathbf{0.10}$ & N/A \\
        Metacontrol & $89.24  \pm 35.57$ & $60.57 \pm 11.17$ & $1.55$        & $0.82$        & N/A \\
        ROSA        & $85.11  \pm 32.48$ & $60.76 \pm 10.29$ & $1.24$        & $1.57$        & N/A \\
        Ours        & $\mathbf{81.52 \pm 19.01}$ & $62.69 \pm 6.06$ & $0.30$ & $0.25$  & N/A \\
        \midrule
        \rowcolor{gray!15}
        \multicolumn{6}{@{}l}{\textbf{SUAVE extended}} \\
        BT          & $94.37  \pm 34.92$ & $20.88 \pm 3.81$  & $\mathbf{0.07}$ & $\mathbf{0.10}$ & $\mathbf{1.09}$ \\
        ROSA        & $92.75  \pm 35.92$ & $18.97 \pm 3.38$  & $1.39$        & $1.67$        & $2.50$ \\
        Ours        & $\mathbf{82.78 \pm 20.68}$ & $\mathbf{21.60 \pm 6.85}$ & $0.36$ & $0.34$ & $1.31$ \\
        \bottomrule
    \end{tabular}
\end{table}

In both SUAVE exemplars, the considered metrics are the time a submerged robot needs to find a pipeline, the distance it manages to inspect along the pipeline, and the reaction time to each introduced fault.
We translate the conditions and resulting actions used in ROSA \cite{silva_rosa_2025} into our DSL and repeat the experiment 100 times with our approach following their evaluation protocol.
\Cref{tab:comparison_sota} shows the results.
For the standard and the extended SUAVE use case, we removed 9 and 7 outliers respectively, i.e., runs in which $t_\text{search} > \text{mean}(t_\text{search}) + 2 \cdot \text{SD}(t_\text{search})$.
This happens, e.g., if the robot did not find the pipeline at all.

\section{Discussion}
\subsection{Research questions}
\begin{table}[ht]
\caption{Evaluation of the ablation study, by varying single parameters as indicated by the leftmost column.}
\label{tab:ablation-study-eval}
\begin{tabular}{c|cc||cccc}
\toprule
\rotatebox{90}{Param.} & \rotatebox{45}{$\text{C}_1^\text{D,C,S}$}& \rotatebox{45}{$\text{C}_2^\text{D,C,S}$}& \rotatebox{45}{$\Delta \frac{s_\text{rslv}}{s_\text{exec}}$} & \rotatebox{45}{$\Delta t_\text{react}$} & \rotatebox{45}{$\Delta t_\text{down}$} & \rotatebox{45}{$\Delta \text{\#rdply}_\text{u} $}  \\
\midrule
\midrule
 \multirow{4}{*}{\rotatebox{90}{Dep. Graph}} & xxx & \checkmark xx & \cellcolor{gray!25}10.67 & \cellcolor{gray!25}-19.77 & \cellcolor{gray!25}5.51 & \cellcolor{gray!25}1.42 \\
  & x\checkmark x & \checkmark \checkmark x & \cellcolor{green!100}86.49 & \cellcolor{gray!25}3.72 & \cellcolor{gray!25}-3.17 & \cellcolor{green!58}-45.64 \\
  & xx\checkmark  & \checkmark x\checkmark  & \cellcolor{gray!25}3.85 & \cellcolor{gray!25}-6.94 & \cellcolor{gray!25}-4.45 & \cellcolor{green!55}-42.86 \\
  & x\checkmark \checkmark  & \checkmark \checkmark \checkmark  & \cellcolor{green!32}27.85 & \cellcolor{gray!25}1.03 & \cellcolor{gray!25}-3.44 & \cellcolor{green!81}-63.27 \\
 \midrule \multirow{4}{*}{\rotatebox{90}{Crit. Lvl.}} & xxx & x\checkmark x & \cellcolor{gray!25}-1.33 & \cellcolor{gray!25}2.28 & \cellcolor{gray!25}9.59 & \cellcolor{gray!25}5.67 \\
  & \checkmark xx & \checkmark \checkmark x & \cellcolor{green!76}66.27 & \cellcolor{red!77}32.23 & \cellcolor{gray!25}0.58 & \cellcolor{green!55}-43.36 \\
  & xx\checkmark  & x\checkmark \checkmark  & \cellcolor{gray!25}1.28 & \cellcolor{gray!25}-9.72 & \cellcolor{gray!25}-1.48 & \cellcolor{gray!25}-12.50 \\
  & \checkmark x\checkmark  & \checkmark \checkmark \checkmark  & \cellcolor{green!28}24.69 & \cellcolor{gray!25}-1.99 & \cellcolor{gray!25}-0.44 & \cellcolor{green!58}-43.75 \\
 \midrule \multirow{4}{*}{\rotatebox{90}{Sys. Impact}} & xxx & xx\checkmark  & \cellcolor{gray!25}4.00 & \cellcolor{gray!25}-17.87 & \cellcolor{gray!25}-3.67 & \cellcolor{green!77}-60.28 \\
  & \checkmark xx & \checkmark x\checkmark  & \cellcolor{gray!25}-2.41 & \cellcolor{gray!25}-4.74 & \cellcolor{gray!25}-12.77 & \cellcolor{green!99}-77.62 \\
  & x\checkmark x & x\checkmark \checkmark  & \cellcolor{gray!25}6.76 & \cellcolor{green!93}-27.51 & \cellcolor{gray!25}-13.41 & \cellcolor{green!86}-67.11 \\
  & \checkmark \checkmark x & \checkmark \checkmark \checkmark  & \cellcolor{red!100}-26.81 & \cellcolor{green!100}-29.39 & \cellcolor{gray!25}-13.65 & \cellcolor{green!100}-77.78 \\
 
\bottomrule
\end{tabular}
\end{table}
The main RQ of this work was to investigate how multiple concurrent failures that might be caused by different faults can be resolved effectively at runtime in robotic systems. 
To answer this question, we proposed a novel approach based on a dependency graph and self-adaptation rules that include criticality levels, each strategy's success probability, and system impact.
To validate whether our approach can handle multiple concurrent failures (RQ1), we evaluated it on SUNSET \cite{wiedholz_sunset_2026}. 
Furthermore, we used this environment to conduct an ablation study that evaluates the specific impact of all components of our proposed system (RQ2).

As shown in Table \ref{tab:ablation-study}, a combination of using dependency graph and criticality levels achieves the highest ratio of resolved failures to executed strategies.
The second-highest ratio is achieved by the combination of all components of our approach.
The fact that these two mean ratios $\frac{f_\text{rslv}}{s_\text{exec}} > 1$ shows, that not only were the correct strategies chosen most of the time, but also that\revtext{, on average, more failures were resolved than strategies were executed. Thus, this implies that at least} some strategies resolved multiple failures at once. 
Thus, we can positively answer \textbf{RQ1} for the SUNSET use case used in our evaluation.
This provides first empirical evidence, indicating that our system fulfills one of our core motivations: resolving multiple concurrent failures by identifying and adapting their root causes.

Furthermore, the full combination of dependency graph, system impact, and criticality levels achieved the lowest number of unnecessary redeploys.
Together with the second highest ratio of resolved failures, this indicates that our full approach provides a conducive configuration for most use cases.
However, if the use case has specific requirements, like a low number of adaptations, our results demonstrate that it can make sense to use subsets of the components in our system.

We show in \Cref{tab:ablation-study-eval} the improvement of adding one component for the planning algorithm to an existing configuration while displaying whether the result has statistical significance.
In this table the configuration $\{D,C,S\}_2$ extends configuration $\{D,C,S\}_1$ by considering the respective parameter. 
\revtext{As in} \Cref{tab:ablation-study}, $\{D,C,S\}$ \revtext{denotes the usage of dependency graph (D), criticality level (C), and system impact (S)}.
The percentage difference of the respective metric, when comparing $\{D,C,S\}_2$ to $\{D,C,S\}_1$, is shown and color-coded to indicate improvement or deterioration.
To report statistical significance, we calculate an ANOVA with post hoc tests, pairwise t-tests and tukey alpha value correction (with a 95\% significance level).
The configuration of the managing system is defined as fixed factor, while the individual metrics are dependent variables.
Values that are not statistically significant are marked with a gray background.
Table \ref{tab:ablation-study-eval} shows that in the performed pairwise comparison of the considered parameters, the number of unnecessary redeploys, the reaction time, and the strategy success rate are influenced with statistical significance.
Especially when adding the system impact to an arbitrary configuration of our approach, the number of unnecessary redeploys decreases significantly compared to the prior configuration.
The opposite effect can be observed regarding the ratio of resolved failures, which decreases when adding system impact to the dependency graph and criticality level enabled.
Both described effects can be explained by Equation \ref{eq:cost}: If system impact is not considered, the strategy with the highest probability of success \revtext{(redeployment in our case)} will be chosen first. 
\revtext{This might lead to the execution of expensive adaptations, even though cheaper ones could have resolved the failure as well, which are prioritized when considering the system impact. 
Consequently, expensive adaptations might be executed later, thus, in our example, reducing the number of unnecessary redeployments.}
The reaction time can also be significantly reduced by considering system impact but only if the criticality level is considered as well.
We see that there is no configuration that significantly reduces system downtime.
This is due to the fact that in every configuration of our managing system and in every combination of detected failures, \ac{ros} nodes will be reactivated or redeployed to resolve this failure.
While the system is operational again after executing this strategy, considering the dependency graph or the system impact significantly decreases the number of redeploys, as described earlier.
Thus, we expect to see significant differences in this metric as well in a scenario in which there is a significant difference between a redeploy and a restart.
An example for this would be the need to establish a connection to hardware components or other calibration sequences that only need to be executed when configuring a component and not also when activating it.
If no dependency graph is used, redeploys of multiple nodes are executed simultaneously, which does not increase the system downtime since the duration of redeploy is roughly equal for all nodes in SUNSET.

The results of our robustness analysis, as seen in \Cref{tab:robustness test}, show that our approach is robust against moderate wrongly specified $P_s$, while adding an update mechanism based on Bayesian learning leads to an ideal estimation of the success probabilities.
The learning component is therefore most valuable in long-running deployments where strategies are executed repeatedly and where initial success rate assumptions may be inaccurate or unavailable. 

To summarize the experiments on SUNSET and answer \textbf{RQ2}: When adaptation costs differ significantly across strategies, system impact scoring should be enabled. 
When all adaptations are roughly equivalent in cost, omitting system impact maximizes the failure resolution ratio. 
In all cases, the dependency graph and criticality level together are the minimum configuration for correct root-cause resolution.

Finally, we compare our system's performance to state-of-the-art self-adaptive robotic approaches for environments with sequential failures (\textbf{RQ3}).
To compare our approach with the existing results reported by \cite{silva_rosa_2025} for ROSA, Metacontrol and the BT baseline, we calculated independent two-sample t-tests. 
We use t-tests without the assumption of equal variances, since we only have the reported means and standard deviations from \textcite{silva_rosa_2025}.
For the mean reaction times, standard deviations were not reported, so we cannot calculate t-tests here.
Our results in \Cref{tab:comparison_sota} show that, in the standard SUAVE scenario, our proposed system outperforms most of the other approaches and competes for the first place against a managing system using a \ac{bt}.
The usage of our proposed system yields the shortest search time and we fall second only to the \ac{bt} approach regarding all other metrics.
In the extended SUAVE use-case our system scores similarly. 
It shows the best search time and comes second to \acp{bt} regarding reaction time.
For inspected distance, however, it yields slightly worse results than ROSA \cite{silva_rosa_2025}.
In the normal SUAVE environment, we did not find significant differences between our approach, BT, Metacontrol and ROSA.
In the extended SUAVE environment, our approach was significantly better than ROSA in search time ($t(159)=-2.4$, $p=0.018$) and distance inspected ($t(145)=3.44$, $p<0.001$).
Compared to BT, our approach showed significantly less search time ($t(161)=-2.85$, $p=0.005$) but no significant difference in distance inspected.

\subsection{Limitations}
As mentioned above, a failure can only be handled if its trigger is known apriori and specified in the \ac{dsl}. 
In reality, this might necessitate an exhaustive fault tree analysis. 
Furthermore, our learning mechanism needs many attempts to estimate the ideal success probability of a strategy for a given failure, and in edge cases such as 100\% it will never converge.

\subsection{Reusability}
\label{sec:reusability}
To encourage others to use and validate our approach on different use cases, we publish all our code \cite{replication_package}. 
Additionally, we provide tooling that makes it easier to configure our system for new use cases: a GUI that automatically generates the file describing the specific rules with the \ac{dsl} and checks the validity of the defined strategies in these rules.
For example, a strategy that adds a node twice to the system without removing it would be invalid, since this will lead to an unsuccessful adaptation of the system by design.

\section{Conclusion \& Future Work}
In this paper, we investigated how a MAPE-K-based managing system can be designed to handle multiple concurrent failures\revtext{, caused by an arbitrary fault, from a set of apriori known ones} in complex robotic systems.
Our evaluation on the SUNSET exemplar shows that for this use case the combination of a dependency graph and criticality-level awareness is the minimum viable configuration for correct failure resolution. 
Together, these two components capture fundamentally different dependency types \,---\, functional dependencies among software components and prioritization dependencies that ensure critical failures are resolved before low-severity ones \,---\, and neither is sufficient alone. 
Considering the system impact adds a tunable layer on top of this foundation, which reduces unnecessary adaptations at the cost of a slightly lower ratio of resolved failures, making it most valuable in systems where adaptation actions differ significantly in their cost and disruption.
We further show, using the SUAVE exemplar, that this generalized capability also works for sequential failure scenarios: our approach achieves the best search time and remains competitive across other metrics against approaches designed primarily for sequential failures.
Taken together, these findings suggest that lightweight, rule-based root-cause analysis \,---\, grounded in a live dependency graph rather than exhaustive fault enumeration \,---\, is a practical and effective foundation for self-adaptive robotic systems.

For future work, our proposed approach can be expanded to (1) resolve failures, even if their resolution is not given explicitly in the adaptation rule set, and (2) learn dependencies between the rules in our \ac{dsl}. 
Furthermore, the validity of these results needs to be evaluated in larger and real-world robotic systems \revtext{compared to} SUNSET and SUAVE.

\section*{Acknowledgments}

The authors would like to thank the Federal Ministry of Economic Affairs and \revtext{Energy} of Germany for funding the described activities through the LuFo VI-3 program, funding code 20F2201D.
%

\printbibliography
\end{document}